\documentclass[11pt]{article}

\usepackage[final]{acl}

\usepackage{times}
\usepackage{latexsym}

\usepackage[T1]{fontenc}

\usepackage[utf8]{inputenc}

\usepackage{microtype}

\usepackage{inconsolata}

\usepackage{graphicx}

\usepackage{amsmath}
\usepackage{amssymb}
\usepackage{amsthm}
\usepackage{mathtools}
\usepackage{booktabs}
\usepackage{multirow}
\usepackage{xcolor}
\usepackage{url}
\usepackage{algorithm}
\usepackage{algpseudocode}
\usepackage{tcolorbox}
\tcbuselibrary{skins,breakable}

\usepackage{enumitem}
\usepackage{bm}

\title{HyperMem: Hypergraph Memory for Long-Term Conversations}

\author{
 \textbf{Juwei Yue\thanks{~Equal contribution.}\textsuperscript{1,2,3}},
 \textbf{Chuanrui Hu\footnotemark[1]\textsuperscript{3}},
 \textbf{Jiawei Sheng\thanks{Corresponding author.}\textsuperscript{1,2}},
 \textbf{Zuyi Zhou\textsuperscript{3}},
 \textbf{Wenyuan Zhang\textsuperscript{1,2}},
\\
 \textbf{Tingwen Liu\textsuperscript{1,2}},
 \textbf{Li Guo\textsuperscript{1,2}},
 \textbf{Yafeng Deng\footnotemark[2]\textsuperscript{3}}
\\
\\
 \textsuperscript{1}Institute of Information Engineering, Chinese Academy of Sciences
\\
 \textsuperscript{2}School of Cyber Security, University of Chinese Academy of Sciences
\\
 \textsuperscript{3}EverMind AI
\\
 \small{
   \{yuejuwei, shengjiawei\}@iie.ac.cn, \{chuanrui.hu, dengyafeng\}@shanda.com
 }
}

\begin{document}
\maketitle

\begin{abstract}
Long-term memory is essential for conversational agents to maintain coherence, track persistent tasks, and provide personalized interactions across extended dialogues.  
However, existing approaches as Retrieval-Augmented Generation (RAG) and graph-based memory mostly rely on pairwise relations, which can hardly capture high-order associations, i.e., joint dependencies among multiple elements, causing fragmented retrieval.
To this end, we propose \textbf{HyperMem}, a hypergraph-based hierarchical memory architecture that explicitly models such associations using hyperedges. 
Particularly, HyperMem structures memory into three levels: \textit{topics}, \textit{episodes}, and \textit{facts}, and groups related episodes and their facts via hyperedges, unifying scattered content into coherent units.
Leveraging this structure, we design a hybrid lexical-semantic index and a coarse-to-fine retrieval strategy, supporting accurate and efficient retrieval of high-order associations.  
Experiments on the LoCoMo benchmark show that HyperMem achieves state-of-the-art performance with 92.73\% LLM-as-a-judge accuracy, demonstrating the effectiveness of HyperMem for long-term conversations.
\end{abstract}

\section{Introduction}

Conversational agents~\cite{MemorySurvey} increasingly serve as long-term companions, requiring coherent multi-hop reasoning, persistent task tracking, and personalized interactions across extended dialogues. 
However, their fixed context windows render historical experiences inaccessible as conversations grow, necessitating effective and efficient long-term memory management~\cite{MemOS, EverMemOS, ExpSeek}.

\begin{figure}[t]
	\centering
    \includegraphics[width=\linewidth]{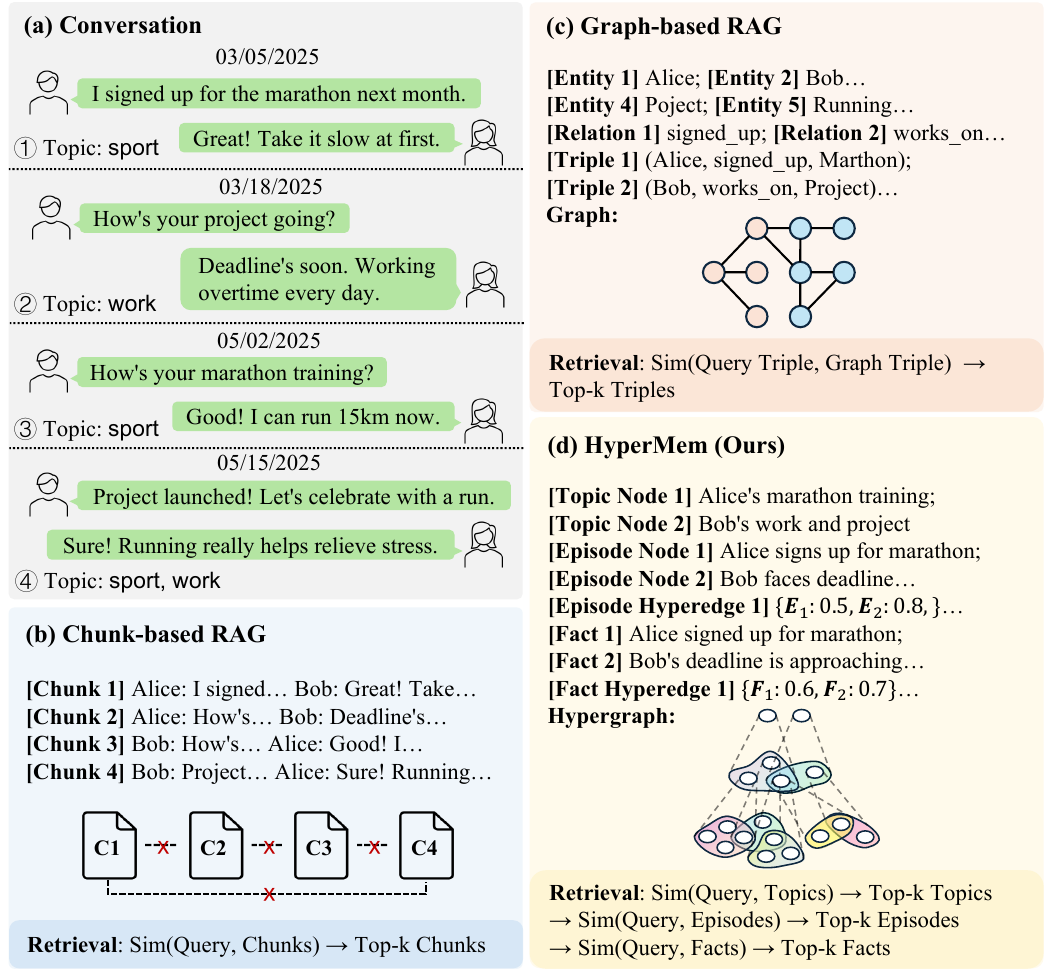}
    \caption{Memory structure comparison across Chunk-based RAG, Graph-based RAG, and our HyperMem.}
    \label{fig:intro}
\end{figure}

Existing approaches such as Retrieval-Augmented Generation (RAG)~\cite{RAGSurvey1, RAGSurvey2} and graph-based memory~\cite{G-Memory, Zep} retrieve external stored related information to enrich the context in response to user queries.
However, both paradigms fundamentally rely on pairwise relationships, which inherently fail to capture \textit{high-order associations}, i.e., joint dependencies among three or more related content elements. 
As shown in Figure~\ref{fig:intro}(a), a conversation may cover multiple topics such as \texttt{sport} and \texttt{work}.
Episodes 1, 3, and 4 are jointly associated under the \texttt{sport} topic and involve multiple facts scattered throughout the dialogue. 
Conventional methods, as shown in Figure~\ref{fig:intro}(b) and (c), can hardly model the holistic coherence among episodes and facts, leading to fragmented retrieval.

To explicitly capture the above high-order associations, we model long-term memory as a hypergraph (Figure~\ref{fig:intro}(d)). 
Unlike conventional graphs with pairwise edges, hypergraphs support \textit{hyperedges} that connect arbitrary node sets, making them uniquely capable of modeling joint dependencies in dialogue. 
Our architecture, namely \textbf{HyperMem}, organizes a three-level memory hierarchy:
(i) \textit{Topic} nodes, representing key conversation themes;
(ii) \textit{Episode} nodes, denoting temporally contiguous dialogue segments centered on a single topic; and (iii) \textit{Fact} nodes, encoding fine-grained details extracted from episodes.
Thereafter, we use hyperedges to explicitly group all episodes sharing the same topic, as well as all facts belonging to the same episode.
These hyperedges may naturally overlap across episodes and facts, reflecting the multifaceted nature of conversational content while preserving semantic coherence within each group.
As a result, semantically scattered information is unified into coherent units, enabling complete and efficient retrieval of high-order associations.

To construct HyperMem, we first detect episode boundaries from the dialogue stream, then aggregate topically related episodes into shared topics using hyperedges, and finally extract fine-grained facts from each episode content.  
For indexing, we leverage lexical cues and exploit dense semantics with hypergraph embedding propagation.
This enables semantically related memories, even if temporally distant, to derive aligned embeddings, thereby facilitating the retrieval of high-order associations.
At retrieval time, HyperMem performs a coarse-to-fine search:
it first identifies relevant topics, then expands to their constituent episodes, and finally selects the most pertinent facts to construct a focused context for response generation.
Our contributions are summarized as follows:
\begin{itemize}[leftmargin=*]
    \item We propose HyperMem, a pioneering three-level hypergraph memory architecture that explicitly models high-order associations via hyperedges, overcoming the limitations of pairwise relation methods to capture holistic coherence.
    \item We leverage the HyperMem structure to derive accurate lexical and semantical indexing, and design a coarse-to-fine retrieval strategy to enable efficient early pruning of irrelevant context.
    \item Experiments on the LoCoMo benchmark achieve state-of-the-art performance with 92.73\% LLM-as-a-judge accuracy, demonstrating the effectiveness of HyperMem for long-term conversations.
\end{itemize}

\section{Related works}

\subsection{Retrieval-Augmented Generation}

RAG has proven effective in mitigating hallucinations~\cite{RAGSystem} and improving reliability~\cite{Self-Reasoning, Self-RAG}, and also serve as a foundation for long-term memory in LLM-powered agents~\cite{HippoRAG, HippoRAG2, REFRAG}.

Vanilla methods retrieve relevant fragments from external sources and use them as context for more grounded responses~\cite{RAG1, RAGRL}.
To enrich relational structures, GraphRAG~\cite{GraphRAG} pioneered knowledge graph construction, inspiring works~\cite{G-Retriever, GRAG, RoG, G-RAG, PathRAG, LightRAG, MiniRAG, SubgraphRAG} that leverage graph topology for structure-aware reasoning and multi-hop retrieval.
For hierarchical modeling, RAPTOR~\cite{RAPTOR}, SiReRAG~\cite{SiReRAG}, and HiRAG~\cite{HiRAG} build tree-structured indices for multi-granular evidence integration.
However, these methods rely on pairwise edges that cannot explicitly group multiple scattered yet semantically related memories.

Recent works~\cite{HyperGraphRAG, Hyper-RAG, OG-RAG, Cog-RAG} preliminarily explore hypergraphs to model multi-entity relations with hyperedges.
However, these approaches are designed for static knowledge bases with determinate corpora, where agentic memory continuously evolves with ongoing dialogues.
Besides, they lack a hierarchical retrieval mechanism capable of preserving semantic coherence across extended dialogues.
Our work pioneers the hypergraph in structuring agentic memory, which has quite different problem settings and technical designs.

\subsection{Memory System of Agents}

Recent agents have used RAG to model long-term memory, where MemoryBank~\cite{MemoryBank}, A-Mem~\cite{A-Mem}, Mem0~\cite{Mem0}, and Zep~\cite{Zep} build structured or graph-based representations for persistence between sessions and tracking of the evolution of facts.
G-Memory~\cite{G-Memory} and LightMem~\cite{LightMem} further explore hierarchical structures and compression for efficiency.

\begin{figure*}[t!]
    \centering
    \includegraphics[width=\textwidth]{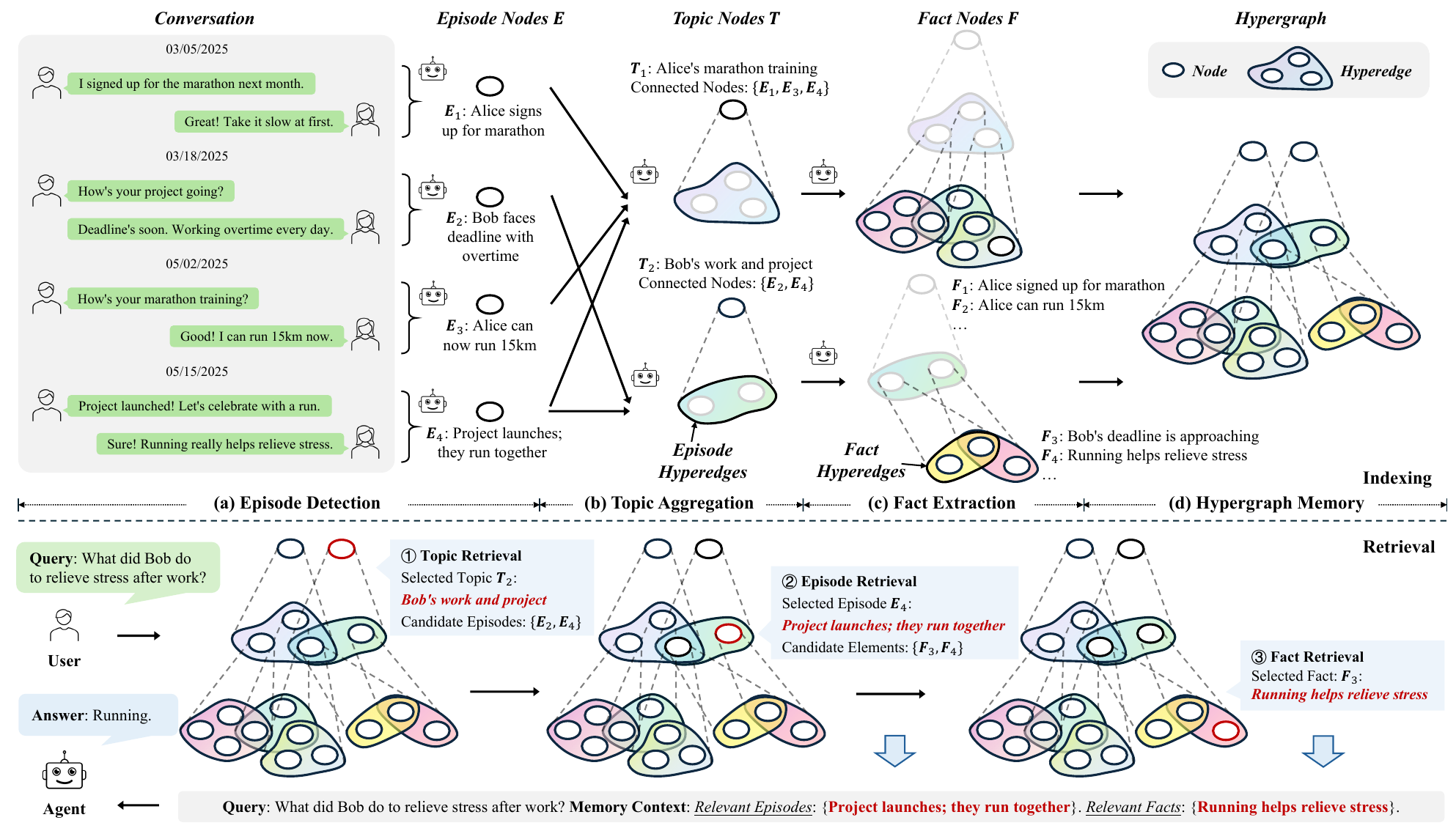}
    \caption{Framework of HyperMem. The indexing detects episode boundaries, aggregates topics via hyperedges, and extracts facts. The retrieval performs coarse-to-fine search from topics to episodes to facts.} 
    \label{fig:framework}
\end{figure*}

In parallel, several approaches eschew explicit retrieval.
MemGPT~\cite{MemGPT} and MemOS~\cite{MemOS} draw on abstractions from operating systems with hierarchical memory and modular scheduling.
MIRIX~\cite{MIRIX} coordinates multi-agent states via shared memory spaces, while Nemori~\cite{Nemori} and MemGen~\cite{MemGen} form compressible or generative latent representations.
MemInsight~\cite{MemInsight}, Mem1~\cite{Mem1}, Memory-R1~\cite{Memory-R1}, and Mem-\(\alpha\)~\cite{Mem-alpha} employ reinforcement learning to autonomously optimize memory storage and retrieval policies.
In contrast, HyperMem explicitly groups topically related memories via hyperedges and employs topic-guided hierarchical retrieval to ensure relevance across temporal gaps.

\section{Approach}

In this section, we present the HyperMem architecture for long-term conversational agents, including hypergraph memory structure, hypergraph construction from dialogue streams, and hypergraph-guided retrieval for response generation.

\subsection{Hypergraph Memory Structure}

To capture higher-order associations among related elements, we model memories with hypergraphs.
Unlike conventional graphs limited to pairwise relations, hypergraphs connect multiple nodes via a single hyperedge.
This enables richer relational modeling and naturally reflects the associative nature of human memory~\cite{anderson2014human}.

To effectively organize this memory, we design a three-level hypergraph architecture, where hyperedges link nodes within each level:
\begin{itemize}[leftmargin=*]
\item \textbf{Topic-level}: Captures dialogues sharing a common theme across long-term interactions, facilitating long-range topical associations.
\item \textbf{Episode-level}: Represents temporally contiguous dialogue segments that describe a coherent event or sub-conversation.
\item \textbf{Fact-level}: Encodes atomic facts extracted from episodes, serving as precise retrieval targets for query-based access.
\end{itemize}
Formally, given an input dialogue stream \(X=\{x_t\}_{t=1}^{T}\), we construct the memory hypergraph as:
\begin{equation}
    \mathcal{H} = (\mathcal{V}^T \cup \mathcal{V}^E \cup \mathcal{V}^F,\ \mathcal{E}^E \cup \mathcal{E}^F),
\end{equation}
where \(\mathcal{V}^T, \mathcal{V}^E, \mathcal{V}^F\) denote the topic, episode, and fact nodes, respectively.
Here, hyperedges \(\mathcal{E}^E\) connect all episode nodes within the same topic along with each node weight \({w}^E \in [0,1]\), while hyperedges \(\mathcal{E}^F\) connect all fact nodes belonging to the same episode with the node weight \({w}^F \in [0,1]\).

\subsection{Hypergraph Memory Construction}

To construct the hypergraph memory, we employ a three-stage process.
We first detect \textit{episodes} by segmenting the raw dialogue stream into atomic sequences, then aggregate topically related episodes into \textit{topics}, and finally extract queryable informative \textit{facts} grounded in their context.

\subsubsection{Episode Detection}
A dialogue stream often interweaves multiple events and shifts topics over time.
Storing it as a monolithic block would obscure event boundaries and entangle events of interest with irrelevant context.  
To address this, we introduce \textbf{\textit{Episode}} to enable precise event boundary preservation and isolate irrelevant content from dialogue context.

\paragraph{Method.} 
To derive episodes, we design an LLM-driven streaming boundary detection mechanism.
Consider an incoming dialogue stream \(X = \{x_t\}_{t=1}^{T}\).
We employ a buffer \(\mathcal{H}\) to pend the history, and determine if the incoming dialogue completes a coherent episode.
Specifically, for each incoming \(x_t\), we add it to \(\mathcal{H}_{<t}\) and invoke an LLM-based boundary detector that evaluates:
(1) \textit{semantic completeness} of current buffer \(\mathcal{H}_{\leq t}\), (2) the \textit{time gap} between consecutive dialogues, and (3) \textit{linguistic signals} indicating topic transition or completion. 

The detector outputs two signals: \texttt{should\_end}, i.e., the buffer forms a semantically complete event, and \texttt{should\_wait}, i.e., the event is still unfolding and requires further input. 
If \texttt{should\_end} is triggered, we create an informative \textit{\textbf{Episode node}}, i.e., \(v^E = (v^E_{\textrm{dialogue}}, v^E_{\textrm{title}}, v^E_{\textrm{episode}})\), where $v^E_{\textrm{dialogue}}$ stores the raw conversation turns, $v^E_{\textrm{title}}$ abstracts a concise subject, and $v^E_{\textrm{episode}}$ offers a brief narrative summary.
The buffer is then cleared, and processing continues with subsequent dialogues.
For the algorithm and prompt, see Algo.~\ref{alg:construction} and Figure~\ref{fig:prompt_stage1}.

\paragraph{Remark.} 
In this way, we process dialogue streams incrementally and segment them into semantically coherent memory units.
This reduces irrelevant context and also improves the convenience of topic organization and retrieval.

\subsubsection{Topic Aggregation}

Episodes capture event-level fragments within contiguous temporal windows.
However, as shown in Figure~\ref{fig:intro}, real-world narratives about a specific topic can also be temporally dispersed.
Existing designs~\cite{Mem0, HyperGraphRAG} usually isolate such correlated associations, making it difficult to retrieve the full narrative.
To address this, we devise \textit{\textbf{Topic}} to aggregate scattered episodes, and leverage hyperedges to connect multiple episodes that belong to the same topic.

\paragraph{Method.} 
Practically, we design an LLM-driven streaming topic aggregation mechanism.
Given the current target episode \(v^E_{\text{cur}}\), we retrieve historical similar episodes \(\mathcal{C}^E\) using lexical and semantic similarity (detailed in \S~\ref{Sec:indexing}). 
By comparing \(v^E_{\text{cur}}\) with \(\mathcal{C}^E\), there are three cases to handle: 
\begin{enumerate}[leftmargin=*]
    \item \textbf{Topic Initialization.} If \(\mathcal{C}^E = \emptyset\), we create a new topic \(v^T = (v^T_{\textrm{title}}, v^T_{\textrm{summary}})\) for \(v^E_{\text{cur}}\).
    Here, $v^T_{\textrm{title}}$ and $v^T_{\textrm{summary}}$ are the title and summary according to \(v^E_{\text{cur}}\) generated by the LLM.
    
    \item \textbf{Topic Creation.} If \(\mathcal{C}^E \neq \emptyset\) but the potential topic of \(v^E_{\text{cur}}\) is different from the existing topics of episodes in \(\mathcal{C}^E\), we create a new topic \(v^T = (v^T_{\textrm{title}}, v^T_{\textrm{summary}})\) for \(v^E_{\text{cur}}\), by comparing \(v^E_{\text{cur}}\) with all episodes in \(\mathcal{C}^E\) by the LLM.
    
    \item \textbf{Topic Update.} If \(\mathcal{C}^E \neq \emptyset\) and the potential topic of \(v^E_{\text{cur}}\) existed in \(\mathcal{C}^E\), we update each matched topic incorporating \(v^E_{\text{cur}}\) and regenerating its metadata \(v^T = (v^T_{\textrm{title}}, v^T_{\textrm{summary}})\).
\end{enumerate}
After this process, we construct a hyperedge \(e^E_t \in \mathcal{E}^E\) linking the topic to all its constituent episodes, and the LLM assigns an importance weight \(w^E_{e,v} \in [0,1]\) to each episode based on its contribution to the topic. 
For the algorithm and prompt, see Algo.~\ref{alg:construction} and Figure~\ref{fig:prompt_stage2}.

\paragraph{Remark.} 
In this way, the resulting topic nodes act as semantic anchors of episodes potentially spanning weeks or months.
This also enables comprehensive retrieval of entire narratives by query matching, regardless of temporal fragmentation.

\subsubsection{Fact Extraction}
Episodes preserve rich narrative context but often contain verbose dialogue that is inefficient for direct query answering. 
To enable query-oriented retrieval, we extract \textbf{\textit{Facts}} with language expressions, the compact assertion grounded in episode context, as fine-grained memory units. 

\paragraph{Method.} 
Given a topic \(t\) and its associated episodes \(\mathcal{V}^E_t\), we use an LLM to identify salient factual assertions, using the full topical context to avoid redundant or trivial extractions. 
Here, each fact node is formed as \(v^F = (v^F_{\textrm{content}}, v^F_{\textrm{potential}}, v^F_{\textrm{keywords}})\), where $v^F_{\textrm{content}}$ records the factual assertion, $v^F_{\textrm{potential}}$ lists query patterns this fact is likely to answer, enabling proactive alignment with user's potential intents, and $v^F_{\textrm{keywords}}$ captures representative terms to facilitate keyword-based retrieval.
To maintain provenance, each fact is explicitly anchored to the original episode(s). 
For each episode \(v^E\), we construct a fact hyperedge \(e^F \in \mathcal{E}^F\) that connects all the facts involved, with the LLM assigning an importance weight \(w^F_{e,v} \in [0,1]\) to reflect the relative importance of each fact. 
For the algorithm and prompt, see Algo.~\ref{alg:construction} and Figure~\ref{fig:prompt_stage3}.

\paragraph{Remark.} 
In this way, the resulting fact nodes serve as atomic query-targeted units.
Unlike raw dialogue for retrieval, $v^F_{\textrm{potential}}$ anticipates relevant queries while $v^F_{\textrm{keywords}}$ supports lexical search, allowing retrieval with concise, directly answerable evidence rather than verbose transcripts.

\subsection{Hypergraph Memory Retrieval}

To respond to the user's query, the agent retrieves relevant memories through a coarse-to-fine process that traverses from \textit{topic} to \textit{episode} to \textit{fact}. 
This combines an offline indexing phase with an online retrieval strategy for practical usage.

\subsubsection{Offline Index Construction}\label{Sec:indexing}
User queries often exhibit both lexical cues and semantic intent, which are crucial to accurately retrieve relevant memories.
To fully leverage both signals, we construct dual indices for all node types, including {topic}, {episode} and {fact}:  
a sparse keyword-based index using BM25~\citep{BM25}, and a dense semantic index powered by Qwen3-Embedding-4B~\citep{Qwen3-emb-rerank}.  
Specifically, each node is first converted into a textual document for BM25 indexing to support exact keyword matching, and then encoded into a dense vector via the embedding model to capture deeper semantic similarity.

\paragraph{Hypergraph Embedding Propagation.}
The nodes linked by the same hyperedge share a common topical context, and are expected to acquire similar representations.
To this end, we propose a lightweight embedding propagation process that enriches node embeddings by aggregating information from their incident hyperedges.
First, we compute a hyperedge embedding as a weighted aggregation of its constituent node embeddings:
\begin{equation}
    \begin{split}
        \bm{h}_e &= \sum_{v \in \mathcal{V}(e)} \alpha_{e,v} \, \bm{h}_v, \\
        \alpha_{e,v} &= \frac{\exp(w_{e,v})}{\sum_{u \in \mathcal{V}(e)} \exp(w_{e,u})},
    \end{split}
\end{equation}
where $\bm{h}_v$ denotes the initial (dense) embedding of node $v$, and $w_{e,v} \in [0,1]$ is the importance weight assigned during topic aggregation, e.g., by an LLM based on narrative contribution.

Next, we refine the representation of each node by aggregating the embeddings of all hyperedges in which it participates:
\begin{equation}
    \bm{h}'_v = \bm{h}_v + \lambda \cdot \operatorname{Agg}_{e \in \mathcal{N}(v)}(\bm{h}_e),
\end{equation}
where $\mathcal{N}(v)$ denotes the set of hyperedges incident to $v$, $\lambda \geq 0$ is a hyperparameter to control the strength of propagation, and $\operatorname{Agg}$ is an aggregation function, e.g., summation. 
See Algo.~\ref{alg:index} for the algorithm.

\paragraph{Remark.} 
This propagation mechanism is inspired by hypergraph neural networks~\cite{feng2019hypergraph}, yet remains lightweight without large-scale fine-tuning.
Empirical studies demonstrate its effectiveness.
Besides, it enables semantically related memories to acquire aligned embeddings, which derive more informative embeddings and also facilitate high-order associations during retrieval.

\subsubsection{Online Retrieval Strategy}

Given a user query \(q\), retrieval proceeds as a structured coarse-to-fine traversal with progressive top-\(k\) selection at each level.

\paragraph{Stage 1: Topic Retrieval.}
We retrieve from the topic-level to establish the topical context.
All topic nodes \(\mathcal{V}^T\) are scored using both keyword and vector indices, with rankings fused via Reciprocal Rank Fusion (RRF):
\begin{equation}
    \mathrm{RRF}(d) = \sum^M_{m=1} \frac{1}{k + \mathrm{rank}_m(d)}
\end{equation}
where \(m\) indexes individual rankers and \(k\) is a smoothing constant.
The RRF-ranked candidates are then refined by a reranker model, which computes fine-grained query-document relevance scores to improve ranking precision.
We select the top-\(k^T\) topic nodes as candidates, which filters out most irrelevant topical contexts.

\paragraph{Stage 2: Episode Retrieval.}
For each selected topic \(t\), we expand to its constituent episodes \(\mathcal{V}^E_t\) via the episode-hyperedge \(e^E_t\).
Following Stage 1, the expanded episodes are scored via RRF and then refined by the reranker. 
We retain the top-\(k^E\) episodes as the results.
This stage ensures that only the query-relevant temporal segments within each topic are preserved.

\paragraph{Stage 3: Fact Retrieval.}
Finally, each retained episode \(e\) is expanded to its supporting facts \(\mathcal{V}^F_e\) through the fact hyperedge \(e^F_e\).
Following the same RRF-then-rerank pipeline, we select the top-\(k^F\) facts as the final retrieval result.

\paragraph{Final Response Generation.}
Instead of using verbose raw dialogue text, we construct the \textit{response context} from the \texttt{content} fields of retrieved \textit{facts}, optionally augmented with the \texttt{summary} fields of their sourced upper-level \textit{episodes} for narrative context.
This design significantly reduces token consumption while preserving answerable information.
The constructed response context is input into the conversational agent, and the response is returned as the answer to the user query.
See Algo.~\ref{alg:retrieval} for the algorithm.

\section{Experiments}

In this section, we conduct experiments to evaluate the effectiveness of our HyperMem.

\begin{table*}[t]
    \centering
    \small
    \begin{tabular}{lccccc}
        \toprule
        \textbf{Methods} & \textbf{Single-hop} & \textbf{Multi-hop} & \textbf{Temporal} & \textbf{Open Domain} & \textbf{Overall} \\
        \midrule
        GraphRAG~\cite{GraphRAG}           
        & 79.55 & 54.96 & 50.16 & 58.33 & 67.60 \\
        LightRAG~\cite{LightRAG}           
        & 86.68 & 84.04 & 60.75 & 71.88 & 79.87 \\
        HippoRAG 2~\cite{HippoRAG2}        
        & 86.44 & 75.89 & 78.50 & 66.67 & 81.62 \\
        HyperGraphRAG~\cite{HyperGraphRAG} 
        & 90.61 & 80.85 & 85.36 & 70.83 & 86.49 \\
        \midrule
        OpenAI~\textsuperscript{\ref{OpenAI}}            
        & 63.79 & 42.92 & 21.71 & 62.29 & 52.90 \\
        LangMem~\textsuperscript{\ref{LangMem}}
        & 62.23 & 47.92 & 23.43 & 71.12 & 58.10 \\
        Zep~\cite{Zep}                     
        & 61.70 & 41.35 & 49.31 & \textbf{76.60} & 65.99 \\
        A-Mem~\cite{A-Mem}                 
        & 39.79 & 18.85 & 49.91 & 54.05 & 48.38 \\
        Mem0~\cite{Mem0}                   
        & 67.13 & 51.15 & 55.51 & 72.93 & 66.88 \\
        Mem0\(^g\)~\cite{Mem0}             
        & 65.71 & 47.19 & 58.13 & 75.71 & 68.44 \\
        MIRIX~\cite{MIRIX}~\(^\dag\)                 
        & 85.11 & 83.70 & 88.39 & 65.62 & 85.38 \\
        Memobase~\textsuperscript{\ref{memobase}} 
        & 73.12 & 64.65 & 81.20 & 53.12 & 72.01 \\
        MemU~\textsuperscript{\ref{memu}}                    
        & 66.34 & 63.12 & 27.10 & 50.01 & 56.55 \\
        MemOS~\cite{MemOS}                 
        & 81.09 & 67.49 & 75.18 & 55.90 & 75.80 \\
        \midrule
        \textbf{HyperMem (Ours)} 
        & \textbf{96.08} & \textbf{93.62} & \textbf{89.72} & 70.83 & \textbf{92.73} \\
        \bottomrule
    \end{tabular}
    \caption{Comparison of HyperMem with RAG-based and memory system methods on the LoCoMo benchmark. Evaluation metric is LLM-as-a-judge accuracy (\%) scored by GPT-4o-mini. $\dagger$ indicates results obtained using GPT-4.1-mini. Results for RAG-based methods are reproduced using their official implementations. Results for memory systems are primarily sourced from~\citet{Mem0, MIRIX, MemOS}.}
    \label{tab:locomo}
\end{table*}

\subsection{Experimental Setup}

\paragraph{Benchmark.}

LoCoMo~\cite{LoCoMo} is a benchmark dataset designed to evaluate long-term memory capabilities in conversational AI systems. 
It contains multi-session dialogues spanning several months, with four categories of questions: single-hop (direct fact retrieval), multi-hop (reasoning across multiple dialogue turns), temporal reasoning (time-related queries), and Open Domain (open-ended questions requiring broader context understanding).

\paragraph{Baselines.}

We compare our approach against representative methods from RAG and memory system.
(1) \textbf{RAG} methods: RAG, GraphRAG~\cite{GraphRAG}, LightRAG~\cite{LightRAG}, HippoRAG 2~\cite{HippoRAG2}, and HyperGraphRAG~\cite{HyperGraphRAG}.
(2) \textbf{Memory system} methods: 
OpenAI~\footnote{\url{https://openai.com/zh-Hans-CN/index/memory-and-new-controls-for-chatgpt/}\label{OpenAI}}, 
LangMen~\footnote{\url{https://langchain-ai.github.io/langmem/}\label{LangMem}}, 
Zep~\cite{Zep}, A-Mem~\cite{A-Mem}, Mem0~\cite{Mem0}, Mem-Graph~\cite{Mem0}, MIRIX~\cite{MIRIX}, 
Memobase~\footnote{\url{https://www.memobase.io/blog/ai-memory-benchmark}\label{memobase}}, 
MemU~\footnote{\url{https://memu.pro/}\label{memu}}, MemOS~\cite{MemOS}. 

\paragraph{Implementation Details.}

We implement HyperMem using \texttt{Qwen3-Embedding-4B} for semantic encoding and \texttt{Qwen3-Reranker-4B} for reranking. 
For answer generation, we employ \texttt{GPT-4.1-mini} with chain-of-thought prompting. 
In hierarchical retrieval, we first retrieve \texttt{100} initial candidates, then select top-\texttt{10} Topics, top-\texttt{10} Episodes, and top-\texttt{30} Facts as the final context. 
Node embeddings are updated with a propagation weight \(\lambda=0.5\) to incorporate hyperedge information. 
For evaluation, we use \texttt{GPT-4o-mini} as the LLM judge and report the average scores across \texttt{3} independent runs.

\begin{table}[t]
    \centering
    \small
    \begin{tabular}{lcc}
        \toprule
        \textbf{Configuration} & \textbf{Overall (\%)} & \textbf{$\Delta$} \\
        \midrule
        \textbf{HyperMem} & \textbf{92.66} & -- \\
        \midrule
        \quad w/o FC & 91.75 &  0.91 $\downarrow$ \\
        \quad w/o EC & 88.90 & 3.76 $\downarrow$  \\
        \midrule
        \quad w/o TR & 91.94 & 0.72 $\downarrow$  \\
        \quad w/o TR \& FC & 91.75 & 0.91 $\downarrow$  \\
        \quad w/o TR \& EC & 88.83 & 3.83 $\downarrow$  \\
        \quad w/o TR \& ER & 90.19 & 2.47 $\downarrow$ \\
        \bottomrule
    \end{tabular}
    \caption{Ablation study. FC: Fact Context, EC: Episode Context, TR: Topic Retrieval, ER: Episode Retrieval.}
    \label{tab:ablation}
\end{table}

\begin{figure*}[!t]
    \centering
    \includegraphics[width=0.9\linewidth]{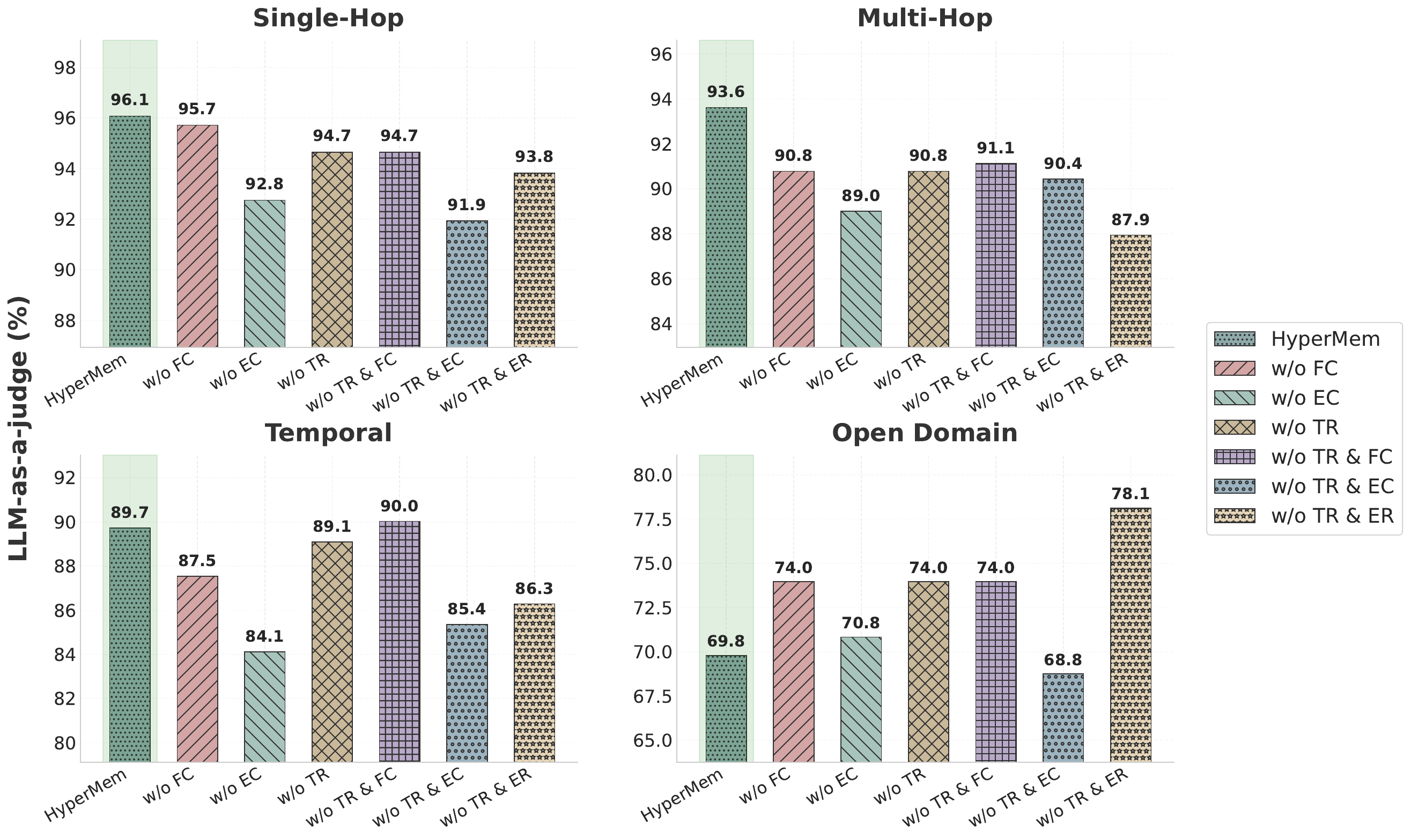}
    \caption{Ablation study across four question categories. FC: Fact context. EC: Episode context. TR: Topic-level retrieval. ER: Episode-level retrieval. The shaded region highlights the full HyperMem configuration.}
    \label{fig:ablation}
\end{figure*}

\subsection{Main Results}

Table~\ref{tab:locomo} presents the main results.
HyperMem achieves the best overall accuracy of 92.73\%, outperforming the strongest RAG method HyperGraphRAG (86.49\%) by 6.24\% and the best memory system MIRIX (85.38\%) by 7.35\%.

Regarding category-wise performance, HyperMem excels on reasoning-intensive tasks.
On Single-hop questions, HyperMem achieves 96.08\%, surpassing HyperGraphRAG by 5.47\%, as the structured fact layer enables precise retrieval of atomic information.
On Multi-hop questions requiring evidence aggregation across multiple dialogue segments, HyperMem reaches 93.62\%, outperforming LightRAG by 9.58\%, demonstrating that hyperedges effectively bind topically related episodes scattered across time for comprehensive evidence collection.
On Temporal questions requiring cross-session reasoning, HyperMem attains 89.72\%, benefiting from the episode layer's preservation of temporal anchors and the hierarchical structure's ability to trace event progression.
Open Domain remains challenging for all methods due to broader knowledge requirements beyond the conversation history.

These improvements stem from two key designs.
Hyperedges explicitly group topically related episodes, ensuring complete evidence retrieval for multi-hop reasoning.
Meanwhile, topic-guided hierarchical retrieval progressively narrows the candidate pool, filtering irrelevant context while preserving temporal coherence.

\subsection{Ablation Study}

As shown in Table~\ref{tab:ablation} and Figure~\ref{fig:ablation}, we conduct ablation study to evaluate the contribution of each component in HyperMem. 
The results reveal that Episode context is the most critical component, as removing it (w/o EC) causes the largest performance drop (-3.76\% overall), particularly affecting Temporal reasoning (-5.61\%). 
The hierarchical retrieval mechanism also proves essential.
Bypassing Topic retrieval (w/o TR) shows moderate impact, but completely flattening the hierarchy to Fact-only retrieval (w/o TR \& ER) significantly degrades Multi-Hop performance (-5.68\%), demonstrating that the hierarchical structure effectively maintains coherent information flow across granularity levels. 
Fact context primarily benefits Multi-Hop reasoning (-2.84\% when removed). 
These findings validate that our three-level memory architecture and hierarchical retrieval strategy work synergistically to achieve optimal performance across diverse question types.

\subsection{Hyperparameter Analysis}

\begin{figure*}[t]
    \includegraphics[width=\textwidth]{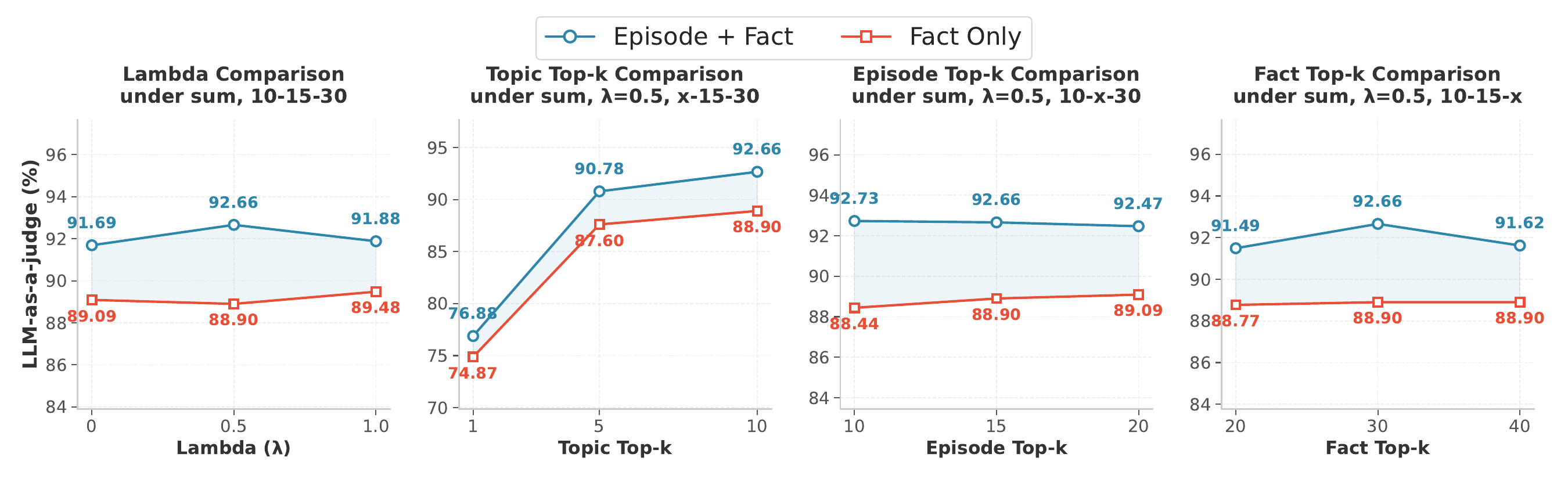}
    \caption{Hyperparameter sensitivity analysis on LoCoMo. We evaluate the impact of embedding fusion weight $\alpha$ and Top-k selection at each hierarchical level (Topic, Episode, Fact) on retrieval performance.}
    \label{fig:hyperparameters}
\end{figure*}

We investigate the sensitivity of HyperMem to key hyperparameters across four dimensions. 
First, the fusion coefficient \(\alpha=0.5\) achieves optimal performance (92.66\%), indicating that balanced integration of semantic similarity and structural retrieval yields the best results.
Second, topic top-k exhibits the most significant impact: increasing from k=1 to k=10 improves accuracy from 76.88\% to 92.66\% (+15.78\%), demonstrating that adequate topical coverage is crucial for capturing relevant context. 
In contrast, episode top-k shows minimal sensitivity (92.73\% at k=10 vs. 92.47\% at k=20), suggesting the system is robust to this parameter. 
Fact top-k peaks at k=30 (92.66\%) with slight degradation at higher values, indicating potential noise introduction from excessive fact retrieval. 
Notably, the ``Fact + Episode'' configuration consistently outperforms ``Fact Only'' by 3-4\% across all settings, further validating the importance of episode-level context in our framework.

\subsection{Efficient Analysis}

\begin{figure}[t]
    \includegraphics[width=\linewidth]{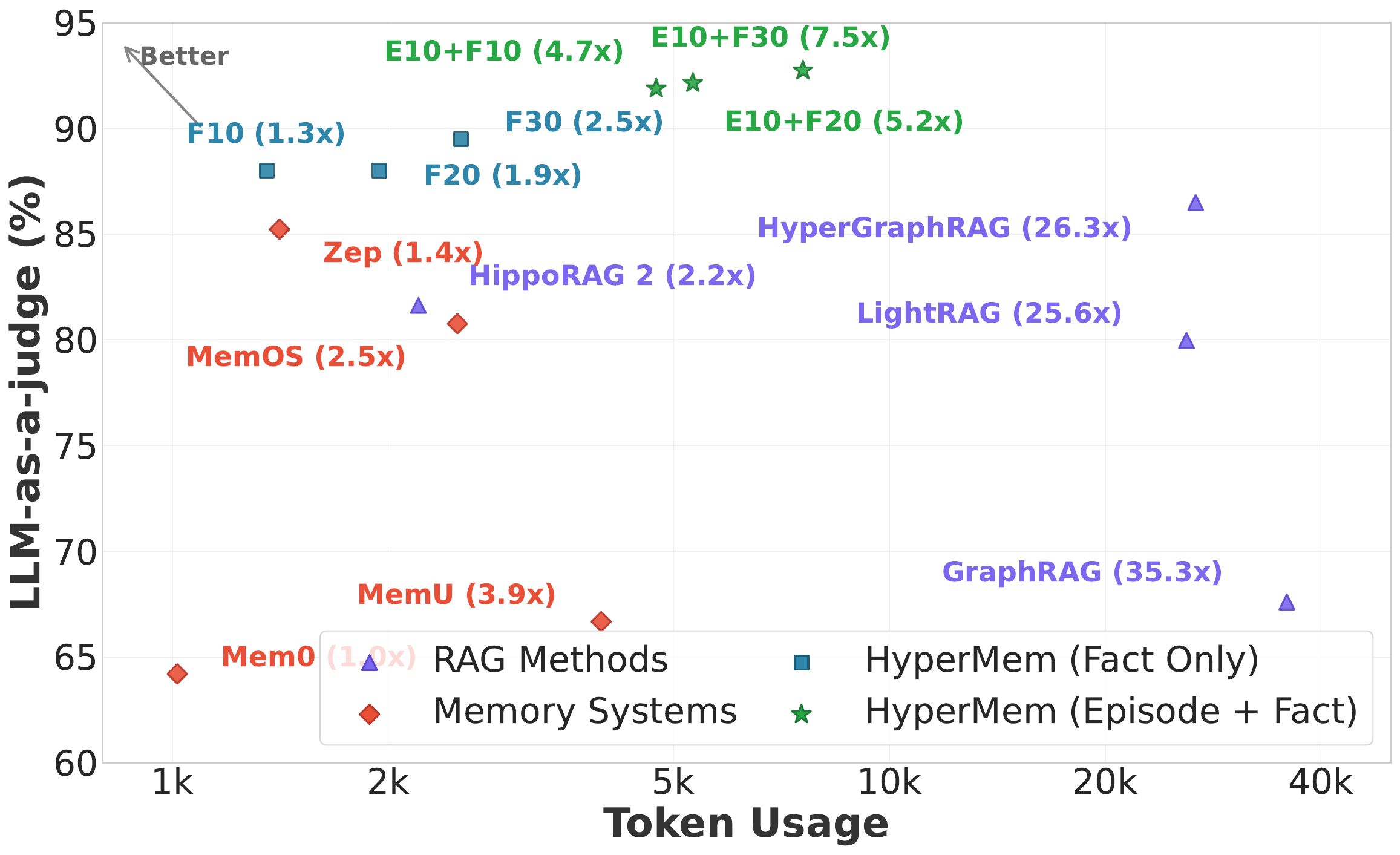}
    \caption{Token usage vs. accuracy comparison. The x-axis shows relative token usage (Mem0 as 1.0$\times$ baseline), and the y-axis shows LLM-as-a-judge accuracy.}
    \label{fig:efficient}
\end{figure}

Figure~\ref{fig:efficient} shows the efficiency-accuracy trade-off.
HyperMem achieves optimal 92.73\% accuracy at 7.5x tokens with the ``Episode + Fact'' configuration, while the ``Fact Only'' configuration already reaches 89.48\% at merely 2.5x tokens, both substantially outperforming RAG-based methods that require 25-35x tokens for lower accuracy (GraphRAG: 67.60\% at 35.3x, HyperGraphRAG: 86.49\% at 26.3x).
The ``Episode + Fact'' configuration consistently outperforms ``Fact Only'' by 3-4\%, demonstrating that episode summaries provide crucial semantic guidance that cannot be compensated by retrieving more facts.

\section{Conclusion}
In this paper, we propose a hypergraph-based agentic memory architecture, namely HyperMem. 
It explicitly models high-order associations among topics, episodes, and facts, overcoming the pairwise limitations of existing RAG and graph-based methods. 
By organizing memory hierarchically and linking related elements via hyperedges, HyperMem unifies scattered dialogue content into coherent units. 
This enables effective lexical-semantic indexing with hypergraph embedding propagation and efficient coarse-to-fine retrieval. 
On the LoCoMo benchmark, HyperMem achieves state-of-the-art 92.73\% LLM-as-a-judge accuracy, demonstrating its strength in long-term conversations.
\footnote{Our source code is about to be released.}

\section*{Limitations}

The current design assumes a single-user scenario, and extending to multi-user or multi-agent settings presents challenges in access control and memory isolation.
Additionally, Open Domain questions remain challenging as they often require external knowledge beyond the conversation history, suggesting opportunities for integrating external knowledge bases.


\bibliography{custom}


\appendix

\section{Algorithm}\label{sec:algorithm}

As shown in Algorithm~\ref{alg:construction}, \ref{alg:index}, and \ref{alg:retrieval}, we provide detailed pseudocode for HyperMem's core procedures. 

\begin{figure*}[t]
\centering
\begin{minipage}[t]{0.495\textwidth}
\begin{algorithm}[H]
    \caption{Hypergraph Memory Construction}
    \label{alg:construction}
    \begin{algorithmic}[1]
    \State \textbf{Input:} Dialogue stream \(X = \{x_t\}_{t=1}^{T}\)
    \State \textbf{Output:} Hypergraph \(\mathcal{H}\)
    \State Initialize \(\mathcal{V}^T, \mathcal{V}^E, \mathcal{V}^F, \mathcal{E}^E, \mathcal{E}^F \gets \emptyset\), buffer \(\mathcal{B} \gets \emptyset\)
    \Statex
    \Statex \Comment{Stage 1: Episode Detection}
    \For{each incoming dialogue \(x_t \in X\)}
        \State \(\mathcal{B} \gets \mathcal{B} \cup \{x_t\}\)
        \State Boundary detection: \((\texttt{end},\! \texttt{wait}) \!\!\gets\!\! \textsc{LLM}(\mathcal{B})\)
        \If{\texttt{end} = \texttt{True}}
            \State \(v^E \gets \textsc{CreateEpisode}(\mathcal{B})\)
            \State \(\mathcal{V}^E \gets \mathcal{V}^E \cup \{v^E\}\), \(\mathcal{B} \gets \emptyset\)
        \EndIf
    \EndFor
    \Statex
    \Statex \Comment{Stage 2: Topic Aggregation}
    \For{each new episode \(v^E_{\mathrm{cur}} \in \mathcal{V}^E\)}
        \State Episode Matching: \(\mathcal{C}^E \gets \textsc{LLM}(v^E_{\mathrm{cur}})\)
        \If{\(\mathcal{C}^E = \emptyset\) (Case 1: Topic Initialization)}
            \State \(v^T \gets \textsc{CreateTopic}(v^E_{\mathrm{cur}})\)
            \State \(\mathcal{V}^T \gets \mathcal{V}^T \cup \{v^T\}\)
        \Else
            \State Topic Matching: \(\mathcal{C}^T \gets \textsc{LLM}(\mathcal{C}^E, v^E_{\mathrm{cur}})\)
            \If{\(\mathcal{C}^T = \emptyset\) (Case 2: New Topic)}
                \State \(v^T \gets \textsc{CreateTopic}(\mathcal{C}^E, v^E_{\mathrm{cur}})\)
                \State \(\mathcal{V}^T \gets \mathcal{V}^T \cup \{v^T\}\)
            \Else \ (Case 3: Topic Update)
                \State \textsc{UpdateTopics}(\(\mathcal{C}^T, v^E_{\mathrm{cur}}\))
            \EndIf
        \EndIf
        \State \(e^E_t \gets (v^T, \textsc{GetEpisodes}(v^T), \bm{w}^E)\)
        \State \(\mathcal{E}^E \gets \mathcal{E}^E \cup \{e^E_t\}\)
    \EndFor
    \Statex
    \Statex \Comment{Stage 3: Fact Extraction}
    \For{each topic \(v^T \in \mathcal{V}^T\)}
        \State \(\mathcal{V}^E_t \gets \textsc{GetEpisodes}(v^T)\)
        \State Fact Extraction: \(\mathcal{F}_t \gets \textsc{LLM}(v^T, \mathcal{V}^E_t)\)
        \For{each fact \(v^F_t \in \mathcal{F}_t\)}
            \State \(\mathcal{V}^F_t \gets \mathcal{V}^F_t \cup \{v^F_t\}\)
            \State Anchor \(v^F_t\) to its source episode(s)
        \EndFor
        \For{each episode \(v^E_t \in \mathcal{V}^E_t\)}
            \State \(e^F_t \gets (v^E_t, \textsc{GetFacts}(v^E_t), \bm{w}^F)\)
            \State \(\mathcal{E}^F_t \gets \mathcal{E}^F_t \cup \{e^F_t\}\)
        \EndFor
    \EndFor
    \Statex
    \State \Return \(\mathcal{H} = (\mathcal{V}^T \cup \mathcal{V}^E \cup \mathcal{V}^F, \mathcal{E}^E \cup \mathcal{E}^F)\)
    \end{algorithmic}
\end{algorithm}
\end{minipage}
\hfill
\begin{minipage}[t]{0.495\textwidth}
\begin{algorithm}[H]
    \caption{Offline Index Construction}
    \label{alg:index}
    \begin{algorithmic}[1]
    \State \textbf{Input:} Hypergraph \(\mathcal{H}\)
    \State \textbf{Output:} Indexed hypergraph with propagated embeddings
    \Statex
    \Statex \Comment{Node Indexing}
    \For{each node \(v \in \mathcal{V}^T \cup \mathcal{V}^E \cup \mathcal{V}^F\)}
        \State Build BM25 and vector index for \(v\)
        \State \(\bm{h}_v \gets \textsc{Encode}(v)\)
    \EndFor
    \Statex
    \Statex \Comment{Hyperedge Embedding}
    \For{each hyperedge \(e \in \mathcal{E}^E \cup \mathcal{E}^F\)}
        \State \(\bm{h}_e \gets \sum_{v \in e} \alpha_{e,v} \bm{h}_v\)
    \EndFor
    \Statex
    \Statex \Comment{Embedding Propagation}
    \For{each node \(v\)}
        \State \(\bm{h}'_v \gets \bm{h}_v + \lambda \cdot \textsc{Agg}_{e \in \mathcal{N}(v)}(\bm{h}_e)\)
    \EndFor
    \end{algorithmic}
\end{algorithm}

\vspace{0.2em}

\begin{algorithm}[H]
    \caption{Online Retrieval Strategy}
    \label{alg:retrieval}
    \begin{algorithmic}[1]
    \State \textbf{Input:} Query \(q\), Indexed \(\mathcal{H}\), Top-\(k\): \((k^T, k^E, k^F)\)
    \State \textbf{Output:} Retrieved context \(\mathcal{R}\)
    \State \(\bm{q} \gets \textsc{Encode}(q)\)
    \Statex
    \Statex \Comment{Stage 1: Topic Retrieval}
    \For{each \(v^T \in \mathcal{V}^T\)}
        \State \(s^T \gets \textsc{RRF}(\textsc{BM25}(q, v^T), \textsc{Cos}(\bm{q}, \bm{h}'_{v^T}))\)
    \EndFor
    \State \(\mathcal{T}_{\mathrm{top}} \gets \textsc{TopK}(\mathcal{V}^T, s^T, k^T)\)
    \Statex
    \Statex \Comment{Stage 2: Episode Retrieval}
    \State \(\mathcal{V}^E_t \gets \bigcup_{t \in \mathcal{T}_{\mathrm{top}}} \textsc{GetEpisodes}(t)\)
    \For{each \(v^E_t \in \mathcal{V}^E_t\)}
        \State \(s^E_t \gets \textsc{RRF}(\textsc{BM25}(q, v^E_t), \textsc{Cos}(\bm{q}, \bm{h}'_{v^E_t}))\)
    \EndFor
    \State \(\mathcal{E}_{\mathrm{top}} \gets \textsc{TopK}(\mathcal{V}^E_t, s^E_t, k^E)\)
    \Statex
    \Statex \Comment{Stage 3: Fact Retrieval}
    \State \(\mathcal{V}^F_{t,e} \gets \bigcup_{e \in \mathcal{E}_{\mathrm{top}}} \textsc{GetFacts}(e)\)
    \For{each \(v^F_{t,e} \in \mathcal{V}^F_{t,e}\)}
        \State \(s^F_{t,e} \gets \textsc{RRF}(\textsc{BM25}(q, v^F_{t,e}), \textsc{Cos}(\bm{q}, \bm{h}'_{v^F_{t,e}}))\)
    \EndFor
    \State \(\mathcal{F}_{\mathrm{top}} \gets \textsc{TopK}(\mathcal{V}^F_{t,e}, s^F_{t,e}, k^F)\)
    \Statex
    \State \Return \(\mathcal{R} \gets \textsc{Compose}(\mathcal{E}_{\mathrm{top}}, \mathcal{F}_{\mathrm{top}})\)
    \end{algorithmic}
\end{algorithm}
\end{minipage}
\end{figure*}

\section{Prompt Templates}\label{sec:prompts}

We present the key prompt templates used in HyperMem. 
Figure~\ref{fig:prompt_stage1} shows the episode boundary detection prompt. 
Figure~\ref{fig:prompt_stage2} describes the topic aggregation prompt for linking related episodes. 
Figure~\ref{fig:prompt_stage3} presents the fact extraction prompt for distilling key information from episodes.

\newtcolorbox{hyperprompt}[1][]{
    enhanced,
    colback=gray!3,
    colframe=black!60,
    coltitle=white,
    fonttitle=\bfseries\small,
    boxrule=0.6pt,
    arc=3pt,
    left=5pt,
    right=5pt,
    top=10pt,
    bottom=3pt,
    attach boxed title to top left={xshift=6pt, yshift=-2mm},
    boxed title style={
        colback=black!60,
        arc=2pt,
        boxrule=0pt,
    },
    #1
}

\begin{figure*}[t]
    \begin{hyperprompt}[title=Episode Detection]
        \textit{You are an episodic memory boundary detection expert. Determine if the newly added dialogue should end the current episode and start a new one.}
        
        \vspace{4pt}
        \textbf{Input:} \hspace{1em} Conversation history: \texttt{\{history\}} \hspace{1em} Time gap info: \texttt{\{time\_gap\}} \hspace{1em} New messages: \texttt{\{new\_messages\}}
        
        \vspace{4pt}
        \textbf{Decision Criteria:}
        \begin{enumerate}[leftmargin=1.5em, itemsep=1pt, topsep=2pt]
            \item \textbf{Substantive Topic Change} (Highest Priority): Do new messages introduce a completely different substantive topic? Is there a shift from one specific event to another distinct event?
            \item \textbf{Intent and Purpose Transition}: Has the fundamental purpose of the conversation changed significantly? Has the core question been fully resolved and a new substantial topic begun?
            \item \textbf{Temporal Signals}: Significant time gap between messages (hours or days)? Long gaps strongly suggest new episodes.
            \item \textbf{Structural Signals}: Clear concluding statements followed by genuinely new topics? Explicit topic transition phrases?
        \end{enumerate}
        
        \vspace{2pt}
        \textbf{Special Rules:} Greetings + Topic = ONE episode; Ignore social formalities and pleasantries; Closures (``Thanks!'', ``Take care!'') stay with current episode.
        
        \vspace{4pt}
        \textbf{Output:} \texttt{\{should\_end: bool, should\_wait: bool, confidence: float, topic\_summary: str\}}
    \end{hyperprompt}
    \caption{Prompt template of episode boundary detection.}
    \label{fig:prompt_stage1}
\end{figure*}

\begin{figure*}[t]    
    \begin{hyperprompt}[title=Topic Aggregation]
        \textit{You are an expert in identifying whether Episodes describe the SAME situation/event/theme. Your task: identify which historical Episodes describe the SAME situation as the new Episode.}
        
        \vspace{4pt}
        \textbf{Input:} \hspace{1em} New Episode: \texttt{\{new\_episode\}} \hspace{1em} Historical Episodes: \texttt{\{history\_episodes\}} \hspace{1em} Existing Topics: \texttt{\{existing\_topics\}}
        
        \vspace{4pt}
        \textbf{Same Situation Criteria} (ALL must be met):
        \begin{enumerate}[leftmargin=1.5em, itemsep=1pt, topsep=2pt]
            \item \textbf{Same Specific Event/Theme}: E.g., ``Jon's career transition'' at different stages. NOT just related topics---``Jon's business'' and ``Gina's business'' are DIFFERENT situations.
            \item \textbf{Narrative Continuity}: Later Episode continues/develops the earlier event. E.g., ``Started X'' $\rightarrow$ ``X encountered problem'' $\rightarrow$ ``X succeeded'' = SAME situation.
            \item \textbf{Identity of Core Subject}: Same specific person's journey, same specific project/initiative, same specific relationship. NOT just same people or same topic category.
            \item \textbf{Temporal Tolerance}: Same situation CAN span multiple time points (weeks or months). Look for recurring discussions or multi-stage developments across time.
        \end{enumerate}
        
        \vspace{2pt}
        \textbf{Aggregation Cases:} $\mathcal{C}^E = \emptyset$ $\Rightarrow$ Create new Topic; \hspace{0.5em} $\mathcal{C}^E \neq \emptyset, \mathcal{C}^T = \emptyset$ $\Rightarrow$ Aggregate into new Topic; \hspace{0.5em} $\mathcal{C}^T \neq \emptyset$ $\Rightarrow$ Update existing Topic.
        
        \vspace{4pt}
        \textbf{Output:} \texttt{\{title: str, summary: str, keywords: list, episode\_weights: dict\}}
    \end{hyperprompt}    
    \caption{Prompt templates of topic aggregation.}
    \label{fig:prompt_stage2}
\end{figure*}

\begin{figure*}[t]    
    \begin{hyperprompt}[title=Fact Extraction]
        \textit{You are an expert in extracting queryable facts from Episodes within a Topic context. Extract atomic, structured facts designed to directly surface answerable evidence.}
        
        \vspace{4pt}
        \textbf{Input:} \hspace{1em} Topic: \texttt{\{topic\}} \hspace{1em} Episodes in this Topic: \texttt{\{episodes\}}
        
        \vspace{4pt}
        \textbf{Extraction Guidelines:}
        \begin{enumerate}[leftmargin=1.5em, itemsep=1pt, topsep=2pt]
            \item \textbf{Answerable Facts}: Focus on facts that directly answer queries, not narrative context. Each Fact should be a standalone, queryable assertion.
            \item \textbf{Provenance}: Maintain explicit links to source Episodes for traceability. Every Fact is anchored to the Episodes from which it originates.
            \item \textbf{Query Anticipation}: Predict potential queries this fact can answer. Store query patterns in the \texttt{potential} field for proactive retrieval alignment.
            \item \textbf{Importance Weights}: Assign salience scores $w \in [0,1]$ based on relevance to the Topic, reflecting each Fact's contribution.
        \end{enumerate}
        
        \vspace{4pt}
        \textbf{Output:} \texttt{\{content: str, potential: str, keywords: list, importance\_weight: float\}}
    \end{hyperprompt}
    \caption{Prompt templates of fact extraction.}
    \label{fig:prompt_stage3}
\end{figure*}

\section{Case Study}

We present four representative cases from the LoCoMo benchmark to illustrate how HyperMem addresses different query types where baselines fail.

\begin{figure*}[h]
    \includegraphics[width=\textwidth]{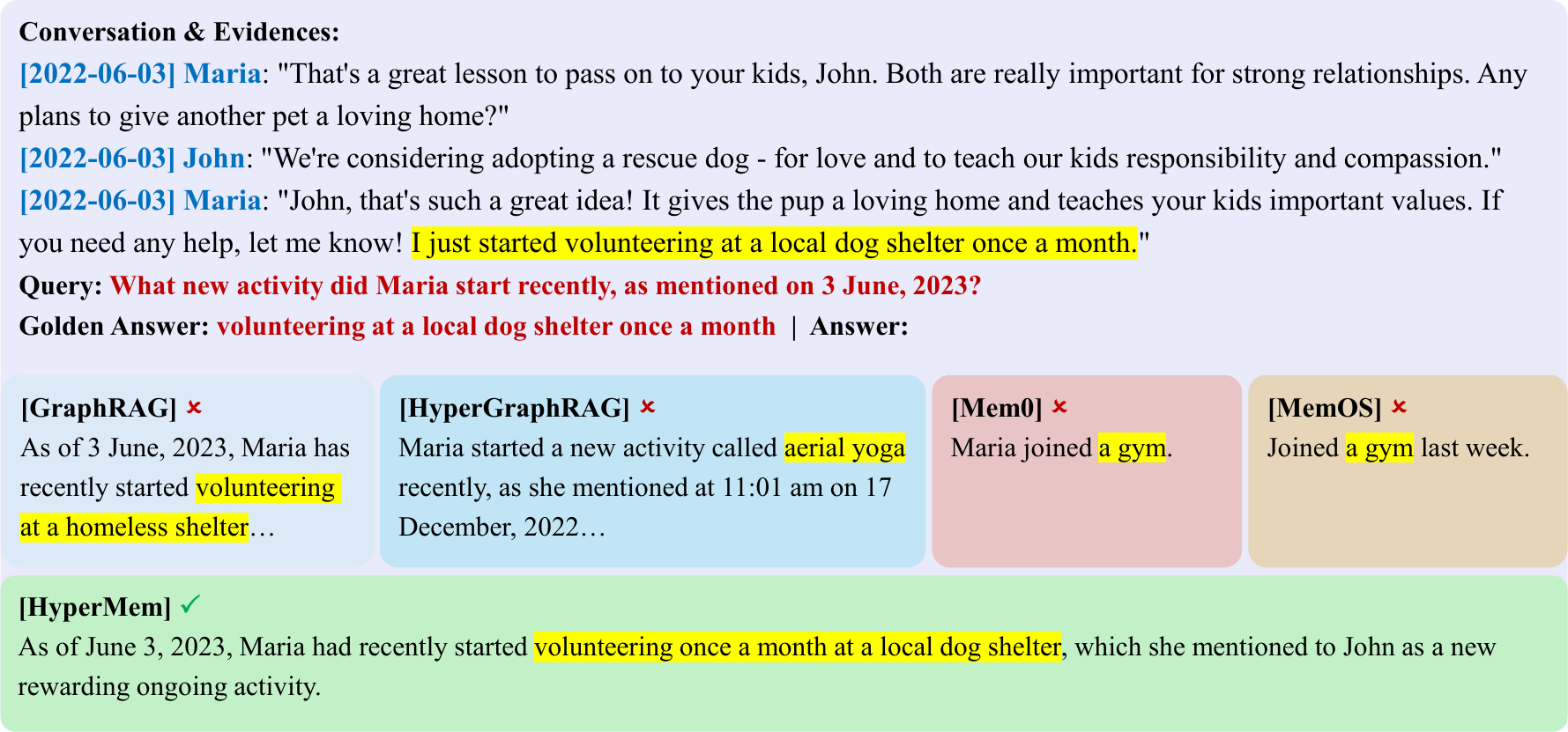}
    \caption{Single-Hop Task. HyperMem precisely retrieves ``dog shelter'' while GraphRAG confuses it with ``homeless shelter.''}
    \label{fig:case_1}
\end{figure*}

\paragraph{Single-Hop Task (Figure~\ref{fig:case_1}).}
This case asks ``What new activity did Maria start recently, as mentioned on 3 June, 2023?'' 
GraphRAG confuses ``dog shelter'' with ``homeless shelter,'' while HyperGraphRAG retrieves ``aerial yoga'' from a different time period. 
HyperMem's hierarchical retrieval navigates through Topic and Episode layers to retrieve the exact Fact containing ``volunteering at a local dog shelter,'' directly matching the golden answer.

\begin{figure*}[h]
    \includegraphics[width=\textwidth]{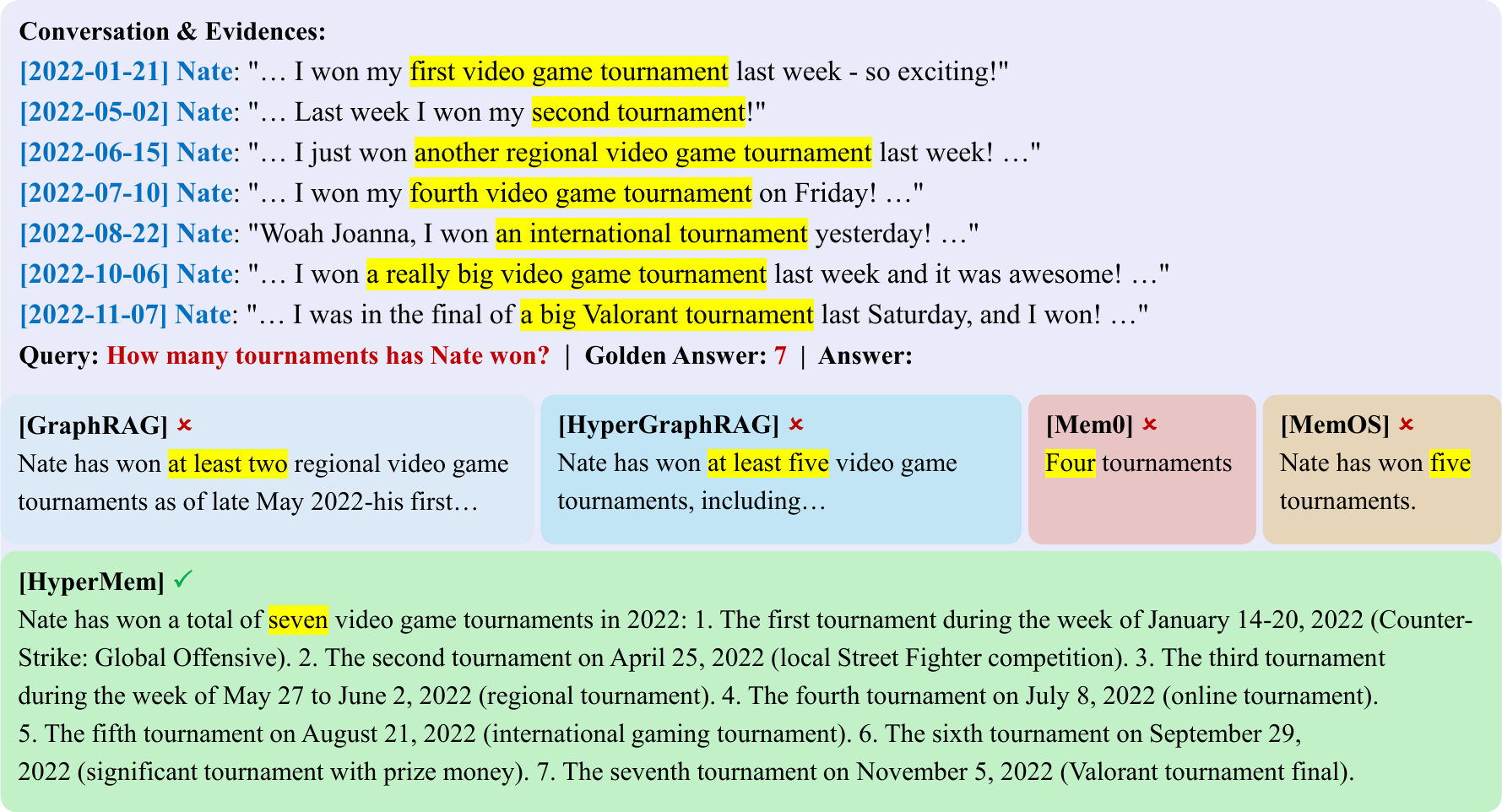}
    \caption{Multi-Hop Task. HyperMem aggregates all 7 tournament mentions across 10 months via Topic hyperedges, while baselines fragment evidence.}
    \label{fig:case_2}
\end{figure*}

\paragraph{Multi-Hop Task (Figure~\ref{fig:case_2}).}
Answering ``How many tournaments has Nate won?'' requires aggregating evidence from 7 sessions spanning 10 months. 
GraphRAG only identifies ``at least two'' due to its pairwise edge structure fragmenting related memories across time. 
HyperMem's Topic hyperedge groups all tournament-related Episodes under a unified thematic anchor, correctly answering ``seven tournaments'' with precise dates for each.

\begin{figure*}[h]
    \includegraphics[width=\textwidth]{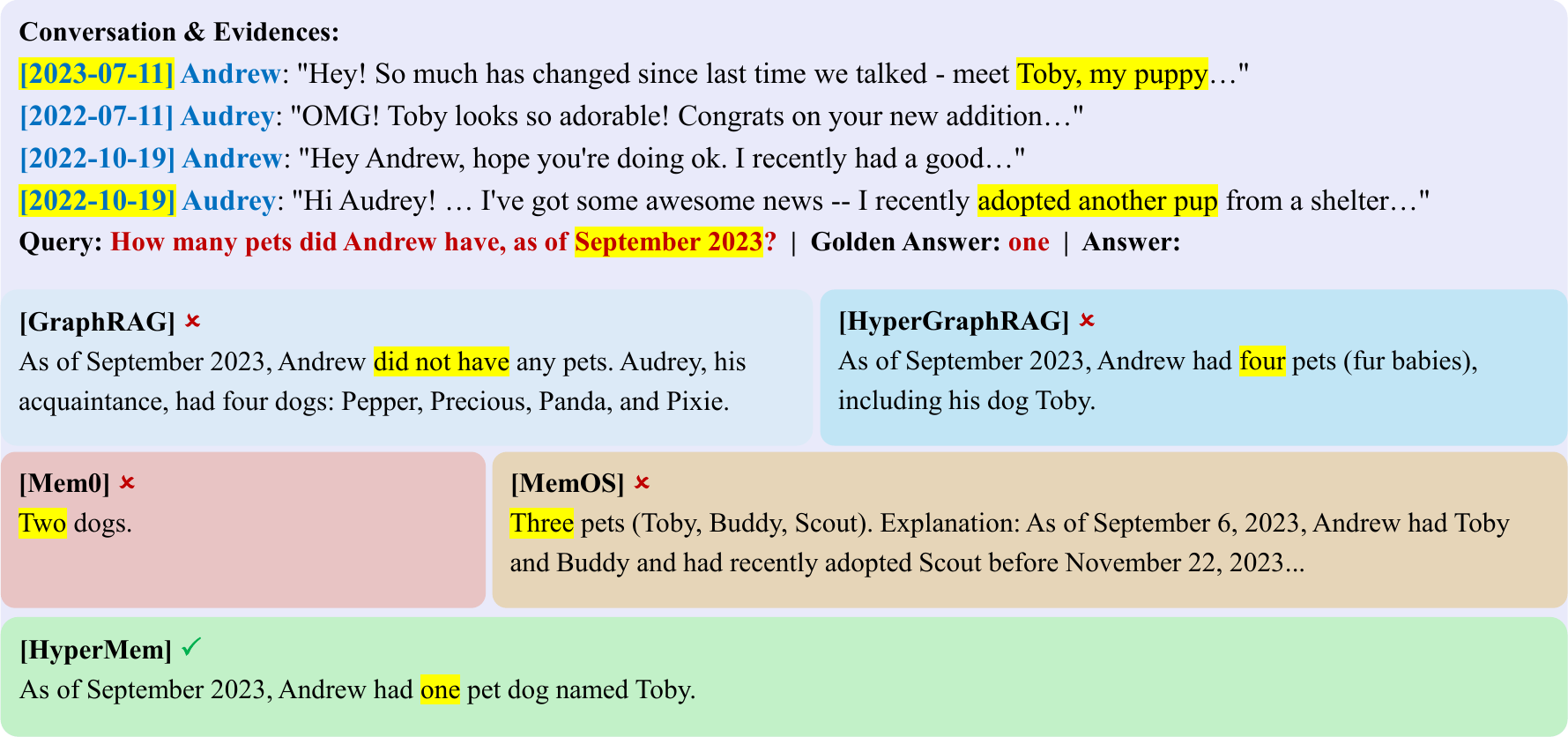}
    \caption{Temporal Reasoning Task. HyperMem correctly identifies one pet at the queried time point, while baselines confuse subjects or miscount.}
    \label{fig:case_3}
\end{figure*}

\paragraph{Temporal Reasoning Task (Figure~\ref{fig:case_3}).}
For the query ``How many pets did Andrew have, as of September 2023?'' 
GraphRAG claims Andrew had no pets by confusing him with another person, while HyperGraphRAG overcounts with ``four pets.'' 
HyperMem correctly answers ``one pet dog named Toby'' because its Episode layer preserves temporal anchors and enables accurate state reconstruction at the queried time point.

\begin{figure*}[h]
    \includegraphics[width=\textwidth]{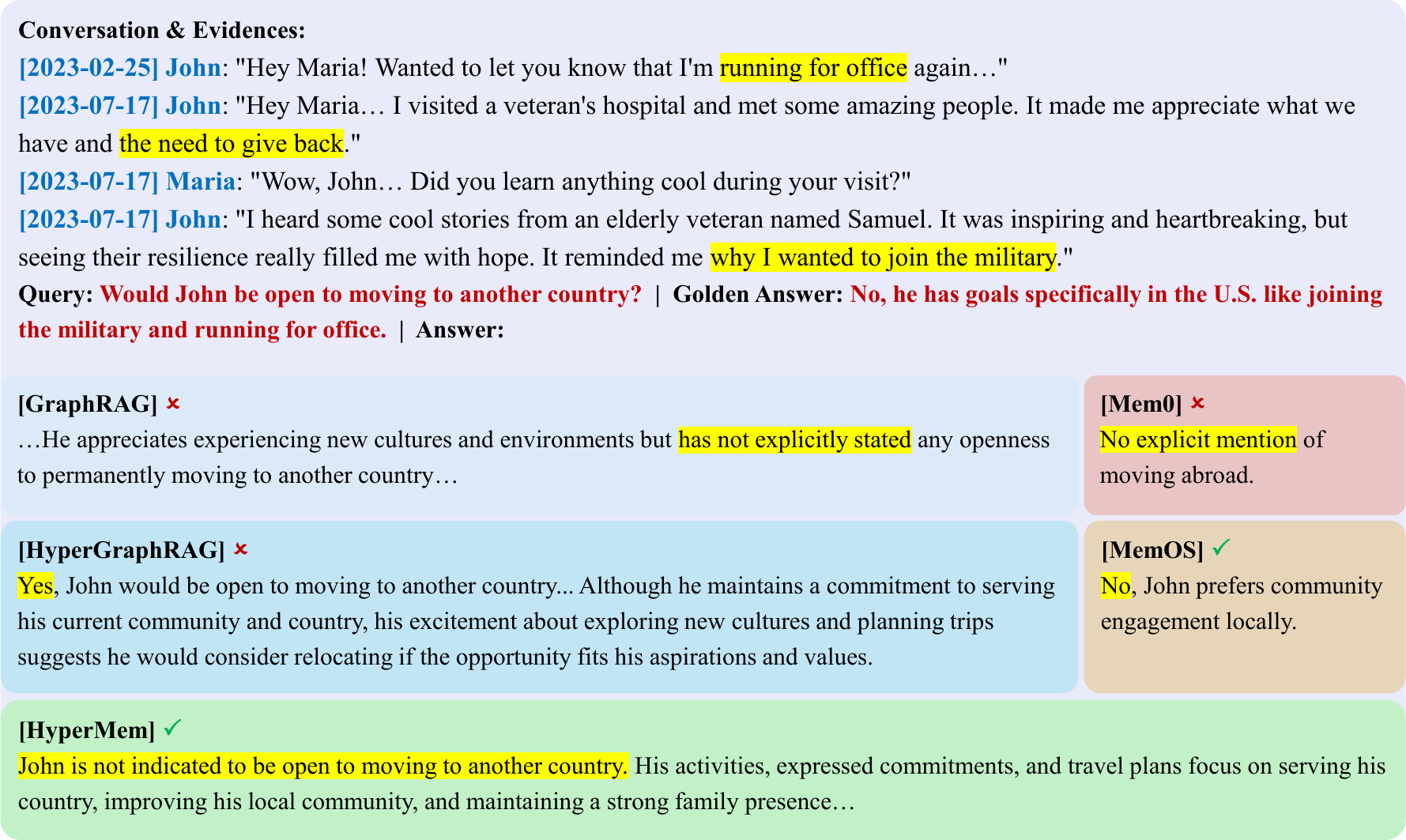}
    \caption{Open Domain Task. HyperMem infers John's commitment to U.S.-based goals, while HyperGraphRAG incorrectly concludes he would relocate.}
    \label{fig:case_4}
\end{figure*}

\paragraph{Open Domain Task (Figure~\ref{fig:case_4}).}
For ``Would John be open to moving to another country?'' 
HyperGraphRAG incorrectly answers ``Yes'' based on superficial travel mentions. 
HyperMem correctly infers ``No'' by synthesizing evidence about John's military aspirations and political campaign goals that anchor him to the U.S. 
The \texttt{potential} field in Fact nodes anticipates such inference patterns.

\end{document}